\documentclass[final,5p,times,twocolumn]{elsarticle}
\usepackage[table]{xcolor}

\usepackage{colortbl}
\usepackage{balance}
\usepackage{amssymb}
\usepackage{array}
\usepackage{longtable}
\usepackage{booktabs}
\usepackage{algorithm}
\newcommand{\Sleep}[1]{\State \textbf{sleep} #1}

\usepackage{algpseudocode}
\usepackage{multirow}
\usepackage{makecell}
\usepackage{threeparttable}
\usepackage{float}
\usepackage{graphicx}
\usepackage{amsthm}
\usepackage{makecell, cellspace, caption}
\usepackage{subcaption}
\usepackage[shortcuts,acronym]{glossaries}
\usepackage{pgfplots}
\usepackage{hyperref}
\DeclareUnicodeCharacter{2212}{−}
\usepgfplotslibrary{groupplots,dateplot}
\usetikzlibrary{patterns,shapes.arrows}
\pgfplotsset{compat=newest}

\newcommand{\FH}[1]{{{#1}}}

\usepackage{booktabs}
\usepackage{tabularx}

\usepackage{dblfloatfix} 
\usepackage{amsthm}
\newtheorem{theorem}{Theorem}
\usepackage[per-mode=symbol,detect-all]{siunitx} 
\DeclareSIUnit{\mbit}{Mbit}
\newcommand{\Mbitps}[1]{\SI{#1}{\mbit\per\second}}
\newcommand{\ms}[1]{\SI{#1}{\milli\second}}
\newcommand{\Hz}[1]{\SI{#1}{\hertz}}
\newcommand{\MHz}[1]{\SI{#1}{\mega\hertz}}
\newcommand{\GHz}[1]{\SI{#1}{\giga\hertz}}
\newcommand{\MB}[1]{\SI{#1}{\mega\byte}}
\newcommand{\m}[1]{\SI{#1}{\meter}}
\newcommand{\km}[1]{\SI{#1}{\kilo\meter}}
\newcommand{\kmh}[1]{\SI{#1}{\kilo\meter\per\hour}}
\newcommand{\pcent}[1]{\SI{#1}{\percent}}
\newcommand{\gcell}[1]{\cellcolor{green!10}#1}
\usepackage[bottom]{footmisc} 

\newacronym{cavs}{CAVs}{Connected and Autonomous Vehicles}
\newacronym{v2x}{V2X}{Vehicle-to-Everything}
\newacronym{cpm}{CPM}{Cooperative Perception Message}
\newacronym{etsi}{ETSI}{The European Telecommunications Standards Institute}
\newacronym{v8}{YOLOv8}{YOLOv8}
\newacronym{bev}{BEV}{Bird's-eye view}
\newacronym{nds}{NDS}{nuScenes Detection Score}
\newacronym{cps}{CPS}{Cooperative Perception System}
\newacronym{v2v}{V2V}{Vehicle-to-Vehicle}

\begin{document}

\begin{frontmatter}
\title{Bandwidth-adaptive Cloud-Assisted 360-Degree 3D Perception for Autonomous Vehicles}
\author[ul-affil]{Faisal Hawlader}\corref{cor}
\cortext[cor]{Corresponding author.}
\author[vassar-affil]{Rui Meireles}
\author[ul-affil]{Gamal Elghazaly}
\author[it-porto-affil]{Ana Aguiar}
\author[ul-affil]{Raphaël Frank}
\address[ul-affil]{Interdisciplinary Centre for Security, Reliability, and Trust (SnT), University of Luxembourg, L-1855, Luxembourg}
\address[vassar-affil]{Computer Science Department, Vassar College, Poughkeepsie, NY 12604, USA}
\address[it-porto-affil]{Instituto de Telecomunicações, Faculdade de Engenharia, Universidade do Porto, Rua Dr. Roberto Frias, 4200–465 Porto, Portugal}
\begin{abstract}
A key challenge for autonomous driving lies in maintaining real-time situational awareness regarding surrounding obstacles 
{\FH{under strict latency constraints}}. 
The high processing requirements coupled with limited onboard computational resources can cause delay issues, particularly in complex urban settings.
To address this, we propose leveraging Vehicle-to-Everything (V2X) communication to partially offload processing to the cloud, where compute resources are abundant, thus reducing overall latency.
Our approach utilizes transformer-based models to fuse multi-camera sensor data into a comprehensive Bird's-Eye View (BEV) representation, enabling accurate 360-degree 3D object detection. 
The computation is dynamically split between the vehicle and the cloud based on the number of layers processed locally and the quantization level of the features. 
To further reduce network load, we apply feature vector clipping and compression prior to transmission.
In a real-world experimental evaluation, our hybrid strategy achieved a \pcent{72} reduction in end-to-end latency compared to a traditional onboard solution. 
To adapt to fluctuating network conditions, we introduce a dynamic optimization algorithm that selects the split point and quantization level 
{\FH{to maximize detection accuracy while satisfying real-time latency constraints}}. 
Trace-based evaluation 
{\FH{under realistic bandwidth variability}} 
shows that this adaptive approach improves accuracy by up to \pcent{20} over static parameterization with the same latency performance.
\end{abstract}
\end{frontmatter}
\section{Introduction}
\label{sec:introduction}
The development of fully autonomous vehicles has driven significant progress in the automotive industry \cite{sonko2024comprehensive, ford_report}, with the potential to transform driving experience by enhancing operational efficiency and safety \cite{bhardwaj2024ai}.
At the core of autonomous driving are perception systems that enable real-time detection of surrounding objects \cite{lee2012perceptual, 10483824}. 
\FH{Such perception capabilities are} critical for navigating complex environments and supporting decision making in motion planning \cite{6866903}.
Industry leaders such as Tesla \cite{marti2019review}, BMW, and Mercedes-Benz face considerable challenges in processing the extensive sensor data (e.g., cameras, radar, and LiDAR) required for 3D object detection \cite{yaqoob2020survey, wang2023distillbev}. 
Recent research has focused on addressing the stringent latency and accuracy requirements associated with autonomous perception tasks \cite{hawlader2024leveraging}. Perception models such as BEVFormer \cite{li2022bevformer} have demonstrated high detection accuracy \cite{Yang2022BEVFormerVA}. However, their computational demands often exceed the capabilities of vehicle hardware \cite{hawlader2024leveraging}, resulting in increased latency and power consumption \cite{gyawali2020challenges}.
For instance, an industrial report published by Ford Motor Company indicated that future vehicles may need to allocate up to \pcent{47} of their energy to onboard computing \cite{ford_report}. 
\FH{These observations highlight a fundamental tension between perception accuracy and real-time latency requirements under constrained onboard computing resources.}

To address onboard computing challenges, researchers have proposed partitioning perception models \cite{mehta2020deepsplit, vnc2023} and offloading computationally intensive layers to the cloud \cite{cohen2020lightweight, choi2018near}. 
While this strategy reduces the onboard computing burden, it also introduces intermediate feature vector transmission latency \cite{torrey2010transfer}, which is problematic for real-time detection with strict latency requirements \cite{mehta2020deepsplit}.
To address the challenge of efficiently transmitting large feature vectors for cloud processing, post-training quantization \cite{liu2021post}, clipping (i.e., outlier removal) \cite{liqun2023clipping}, and compression \cite{hawlader2024leveraging} can be employed. 
\FH{Together, these techniques significantly reduce the amount of transmitted data.}
By reducing bandwidth requirements and transmission latency \cite{wons_2025}, these techniques enable real-time processing within hybrid computing environments \cite{liqun2023clipping}.
However, quantization, clipping, and compression can negatively impact detection quality due to data loss \cite{liqun2023clipping}. 
\FH{Moreover, in practical vehicular deployments, latency is not only determined by computation, but also by fluctuating wireless bandwidth and mobility induced variability. Static offloading configurations may therefore violate latency bounds under changing network conditions.}
Finding an optimal trade-off between end-to-end delay and detection quality \FH{under realistic network dynamics} remains a critical area of research.

To explore these trade-offs, we propose a dynamic hybrid computing strategy based on BEVFormer that integrates cooperative perception for 360-degree 3D detection and adaptively adjusts the split layer and quantization level to optimize perception under real-time constraints.
Cooperative perception enables vehicles and infrastructure to share sensor data via \ac{v2x}, overcoming limitations such as occlusions and sensor range constraints \cite{he2019cooperative}. 
By exchanging \acp{cpm}, which include information about detected objects, this method extends perception beyond onboard sensors \cite{ts2023103}. 
In our approach, the vehicle performs lightweight feature extraction locally while offloading intensive computations to the cloud, combining local and cloud processing to improve real-time performance.
Experimental results showed a \pcent{72} average reduction in end-to-end delay compared to onboard-only computing, for matched quantization levels.

\FH{Different split depths and quantization levels induce distinct latency accuracy trade-off profiles.}
With that in mind, we propose a dynamic parameter selection scheme that, given the available network bandwidth, maximizes detection accuracy while satisfying a target latency bound. Our evaluation demonstrates that the dynamic strategy can improve accuracy, relative to a static parameterization with the same end-to-end latency, by \SI{10} to \pcent{20}, over a wide gamut of network bandwidth and latency bound combinations.
Our contributions can be summarized as:
\begin{itemize}
    \item \textbf{Hybrid-computing perception scheme:} We introduce a BEVFormer-based hybrid computing perception scheme that is able to split computation between vehicle and cloud, transferring data over V2X. It applies data quantization, clipping, and compression, to reduce offloading latency while preserving detection quality.
    \item \textbf{Real-world testing:} We ran tests in a real-world scenario with vehicular mobility and V2X integration. We benchmarked both a fully-onboard computing solution, and our proposed hybrid scheme. We present an analysis of the impact of different hybrid-computing parameterizations on detection accuracy and overall latency.
    \item \textbf{Dynamic parameter selection:} We extend the base hybrid-computing scheme with a constrained-optimization dynamic parameter selection algorithm. 
    It maximizes detection accuracy while \FH{respecting strict latency constraints under} volatile network conditions.   
\end{itemize}

The remainder of this paper is organized as follows: Section \ref{sec:related_work} reviews the relevant literature. Section \ref{sec:methodology} details our proposed method, including the on-board and cloud components, as well as the test route and communication technologies used. 
Section \ref{sec:experiments_results} presents experimental results, focusing on the latency versus accuracy trade-off. 
Section~\ref{sec:dynamic_offloading_parameter_selection} introduces the hybrid-computing parameter optimization scheme. Finally, Section \ref{sec:conclusion_future_work} summarizes our findings and outlines potential future research directions.

%
%
\section{Related Work}
\label{sec:related_work}
Cooperative perception \cite{he2019cooperative} has gained significant attention as a method to enhance the situational awareness of autonomous vehicles \cite{sonko2024comprehensive, gao2024vehicle, wang2025cmp}. 
By enabling vehicles to share sensor data and computational resources \cite{girshick2015fast, bai2025collaborative}, these systems can significantly improve object detection \cite{chiu2025v2v, wang2023distillbev} and prediction in complex environments \cite{10001050}.
Several recent studies have explored vehicle-to-cloud (V2C) communication to offload perception tasks to remote servers \cite{chen2025leveraging, hawlader2024leveraging, yaqoob2020survey}. 
By utilizing the superior compute capabilities of the cloud \cite{wang2025survey}, these approaches reduce the onboard processing burden and energy consumption \cite{VCPKVO_HPCCT22, ford_report}.
Early works in the field, such as \cite{marvasti2020bandwidth, 10522676}, focused on transmitting raw data to the cloud. 

However, these approaches suffered from bandwidth limitations and high transmission latency \cite{ren2015faster, faissal_wons}, making them unsuitable for real-time applications \cite{10184097}. 
To address these challenges, prior works such as \cite{torrey2010transfer, liqun2023clipping} explored feature-level offloading \cite{lee2012perceptual}, where intermediate neural feature vectors are transmitted instead of raw data \cite{10522676}, significantly reducing bandwidth usage \cite{vrevrabek2014comparison}. 
However, the size of these feature vectors can still be prohibitively large \cite{choi2018deep, mehta2020deepsplit}, especially when generated by deep neural networks like ResNet101 \cite{he2016deep} or BEVFormer \cite{li2022bevformer}. 
Recent studies, such as \cite{mehta2020deepsplit}, have proposed various methods for compressing feature vectors \cite{cohen2020lightweight}, including quantization and lossy compression \cite{cohen2020lightweight, liu2020computing}. 
Although these methods reduce the size of the transmission data, they often result in a degradation of the detection accuracy \cite{gyawali2020challenges}. 
\FH{Similarly, approaches such as DistillBEV~\cite{wang2023distillbev} improve computational efficiency through knowledge distillation.} 
Our work builds on these efforts by \FH{incorporating percentile based clipping within a bandwidth-adaptive hybrid offloading framework} that minimizes unnecessary feature data while retaining key information for accurate 3D object detection.
We also evaluate the effectiveness of lossless compression in combination with different floating-point precisions to achieve a balance between latency, bandwidth, and accuracy.

Nevertheless, the performance of feature offloading systems heavily depends on the choice of the split layer \cite{vnc2023, xiao2022perception} that is, the point at which the neural network is divided between the vehicle and the cloud \cite{yaqoob2020survey, babaiyan2025deep}. 
Prior studies have largely relied on static split configurations \cite{cohen2020lightweight, wons_2025}, assuming stable or idealized communication links \cite{sullivan2012overview}.
\FH{For example, DeepSplit~\cite{mehta2020deepsplit} investigates model partitioning strategies to reduce inference latency, typically relying on profiled or fixed communication characteristics. However, such approaches do not explicitly integrate real-time bandwidth estimation into the inference loop.}
However, in real-world V2X deployments \cite{bai2025cellular}, communication links are subject to frequent variations in latency \cite{quah2025data}, jitter, and available bandwidth due to vehicle mobility and fluctuating network load \cite{tavasoli2025data}. 
In such settings, static offloading strategies are vulnerable to latency violations or inefficient resource usage \cite{khan2025dynamic}.
Without real-time adaptability \cite{khan2025dynamic}, such systems risk exceeding latency budgets or underutilizing available resources, ultimately degrading perception accuracy or introducing unnecessary delays.

Additionally, while most prior studies rely on static environments or simulation-based evaluations, our system is validated through real-world vehicular experiments using V2X communication. 
This enables us to account for network jitter and bandwidth variability, providing a more robust evaluation of real-time capabilities and performance of the \ac{cps}.
Furthermore, we introduce a dynamic parameter selection algorithm that adapts the split layer and quantization level, optimizing detection accuracy while satisfying latency constraints under fluctuating network conditions.
%
\section{Methodology}
\label{sec:methodology}
In this section, we describe the system architecture and hybrid-computing methodology for the proposed 360-degree 3D perception framework.
Multi-view (6x) camera images are processed using BEVFormer \cite{li2022bevformer}, a 3D object detection framework for autonomous driving. 
\FH{
The perception pipeline is partitioned between the vehicle and a remote cloud server, where early network layers are executed onboard and deeper layers are offloaded according to the selected split configuration. Intermediate feature representations are transmitted over the wireless link for cloud processing.
}
The model outputs 3D bounding boxes with object positions, orientations, and sizes in a \ac{bev} space, making it well suited for perception. 
After completion of inference under the selected split configuration, the detected 3D objects are then encoded into \acp{cpm} and broadcast to nearby vehicles or infrastructure via standardized V2X communication protocols defined by \ac{etsi} \cite{ts2023103}.
\FH{The hybrid-computing evaluation accounts for realistic vehicular communication effects through empirically measured effective transfer latency, capturing bandwidth variability.}
\begin{figure*}[!t]
  \centering
  \begin{subfigure}[t]{0.3\textwidth}
    \centering
    \includegraphics[width=\linewidth]{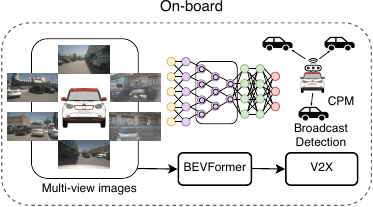}
    \caption{Onboard computing scenario}
  \end{subfigure}~\hspace*{1cm}~\begin{subfigure}[t]{0.6\textwidth}
    \centering
    \includegraphics[width=\linewidth]{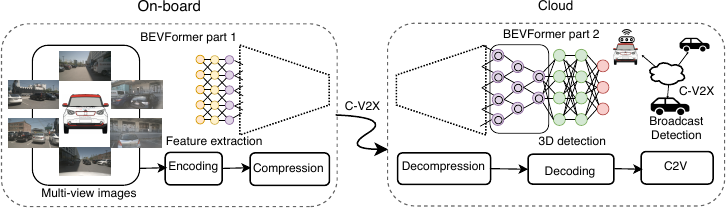}
    \caption{Hybrid computing scenario}
  \end{subfigure}
    \caption{In the Onboard Computing scenario, the BEVFormer model runs locally, transmitting detection results as CPMs over ITS-G5. In the Hybrid Computing scenario, a compressed feature vector is sent via C-V2X to the cloud for intensive processing, with detection results broadcast to nearby vehicles.}
    \label{fig:both_scenarios}
\end{figure*}
\subsection{Test Scenarios \& Routes}
\label{subSec:test_route}
The tests were carried out on public roads in the Kirchberg area of Luxembourg, which \FH{includes various} road layouts and traffic conditions.
This environment allowed us to evaluate perception scenarios in highly realistic settings. 
Communications between vehicles and infrastructure were handled by the YoGoKo Y-Box module, which supports ITS-G5 V2X and C-V2X technologies. 
For more information on the test vehicle, sensors, hardware, and software stack, refer to \cite{robocar}.
For this work, we set up two distinct scenarios, as shown in Figure~\ref{fig:both_scenarios}.

In the \textbf{onboard computing scenario}, multi-view images are fed into the BEVFormer model, with all perception tasks performed locally. The detection results are then encoded into a \ac{cpm} and transmitted to nearby vehicles and infrastructure via ITS-G5.  
The experimental route spanned a distance of approximately \km{1.5}\footnote{\textbf{Onboard computing test route:} \url{http://g-o.lu/3/GsHC}}. 
Within this setup, we evaluated the transmission of \acp{cpm} to assess the reliability and performance of V2X communication in real world conditions.
To ensure consistent measurements, we placed a stationary receiver at specific coordinates. This provided a fixed reference point for evaluating V2X communication quality as the transmitting vehicle moved along the designed test route. 
\FH{The stationary receiver enables controlled and repeatable measurement of CPM latency and packet delivery as a function of distance and channel quality, while avoiding additional variability from dual mobility. Although dense multi-vehicle scenarios may introduce stronger multipath and contention effects, this setup isolates communication performance under single vehicle mobility.}
%
\begin{table}[!b]
\begin{center}
\resizebox{0.85\linewidth}{!}{%
\begin{tabular}{ll}
\textbf{Platform} &  \textbf{Hardware Configuration}\\
\midrule
\multirow{3}{*}{Local ($\approx$30W)}  
    & NVIDIA Jetson Orin\\ 
    & 2048 CUDA Cores, 131.4 TOPS (INT8)\\
    & 8-core ARM Cortex-A78AE\\
\midrule
\multirow{4}{*}{Cloud ($\approx$3000W)}  
    & 2x Intel Xeon Skylake CPUs (56 cores total)\\
    & 4x NVIDIA Tesla V100 GPUs (16 GB each)\\
    & 20480 CUDA Cores, 500 TFLOPS (FP16)\\
\bottomrule
\end{tabular}
}
\caption{Hardware setups for the onboard and cloud computing platforms described in Section~\ref{subSec:test_route}.}
\label{Tab:Methodology_hardware_configuration}
\end{center}
\end{table}

In the \textbf{hybrid computing scenario}, 
\FH{BEVFormer is partitioned into a local component and a cloud component.}
Multi-view images are processed locally by BEVFormer 
\FH{through its early network layers}
to extract feature vectors, which are then clipped, compressed, and sent to the cloud via C-V2X. 
\FH{The cloud server executes the remaining layers to complete 3D object detection.}
The detection results are encoded into \acp{cpm} and broadcast to surrounding vehicles via C-V2X, enabling cooperative perception.
The test route spans approximately \km{4}\footnote{\textbf{Hybrid computing test route:} \url{http://g-o.lu/3/96TS}}. 
We use the cellular mode of C-V2X to communicate with a cloud server located at the University of Luxembourg. 
Twelve commercial base stations, including 4G and 5G (non-standalone) sites, operate along the route on low-band (\MHz{700}) and mid-band (\GHz{3.6}) frequencies. 
{\FH{Drive tests along this route showed}} an average downlink throughput of \Mbitps{57} for 4G and \Mbitps{115} for 5G, with uplink speeds ranging from 25 to \Mbitps{35}. 
\FH{These measurements were obtained under real vehicular mobility, capturing practical bandwidth fluctuations and scheduling variability typical of commercial cellular deployments.}
We use UDP for data offloading to the cloud \FH{to minimize transport-layer overhead and end-to-end latency} \cite{9239945}. 
\FH{Given the real-time constraints of perception offloading, latency minimization is prioritized over strict reliability guarantees.} Potential packet loss is mitigated at the application level through frame-level processing and periodic updates.
\subsection{Hardware Configuration \& Detection Model}
\label{subSec:hardware_configuration}
The hardware configurations used in this study are outlined in Table~\ref{Tab:Methodology_hardware_configuration}. The onboard setup utilizes a Jetson Orin, selected for its low power consumption and processing capabilities in embedded perception tasks. 
The cloud platform employs 4 Tesla V100 GPU nodes, designed to handle computationally intensive tasks. For more details on GPU node configurations, we refer the reader to \cite{VCPKVO_HPCCT22}.
We use the BEVFormer model with a ResNet101 backbone \cite{7780459}, initialized from the FCOS3D checkpoint \cite{Yang2022BEVFormerVA}. 
BEVFormer is a transformer-based 3D object detection framework designed for autonomous driving perception.
It processes multi-view camera images to produce a \ac{bev} representation of the scene, enabling 360-degree perception. 
The model architecture consists of three key stages: (i) a CNN-based backbone (e.g., ResNet101) extracts 2D features from multi-view camera inputs, (ii) a view transformer fuses the features into a unified \ac{bev} space, and (iii) a \ac{bev} encoder refines these representations using temporal and spatial self-attention. 

The detection head then predicts 3D bounding boxes, including object positions, orientations, and sizes. 
Although BEVFormer is trained using LiDAR data for supervision \cite{9607436}, it operates solely on camera inputs during inference, making it well suited for real-time, camera-only 3D perception. 
For more details on BEVFormer, we refer the reader to \cite{li2022bevformer}.
In our hybrid setup, we split computation after the initial backbone layers, performing feature extraction onboard. 
The remaining backbone layers, view transformation, \ac{bev} encoding, and 3D detection are offloaded to the cloud for efficient processing.
\FH{The split point is selected to balance onboard computational load and transmitted feature dimensionality, directly impacting $lat_{local}$ and communication latency as formalized in Eq.~\ref{eq:latency_total}. Alternative split positions are explored experimentally in Section~\ref{sec:experiments_results}.}
%
%
%
\subsection{Input Dataset and Evaluation Metric}
\label{SubSec:dataset_evaluation}
For this study, we use the \textbf{nuScenes} dataset \cite{9607436}, a large-scale dataset specifically created for autonomous driving research. 
The dataset consists of 1600$\times$900 resolution images from six cameras, five radars, and one LiDAR, providing full 360-degree coverage, perfectly aligning with our objective of achieving 3D detection.
The dataset also includes detailed annotations for 3D object detection, tracking, and segmentation across various classes such as vehicles, pedestrians, and cyclists.
We use a BEVFormer model pre-trained on the nuScenes dataset without performing any additional training.
\FH{The dataset is used exclusively for perception accuracy evaluation under different quantization, clipping, and compression configurations, while communication performance is evaluated separately through real-world experiments and trace-based analysis.}
The evaluation aims to quantify potential detection accuracy loss resulting from quantization, clipping, and compression.
To evaluate performance across all cases, we use the \textbf{\ac{nds}}, which offers a comprehensive assessment of detection tasks. The \ac{nds} ranges from 0 to 1, is calculated using the following formula \cite{li2022bevformer}:

\begin{equation}
NDS = \frac{1}{10} \left( 5 mAP + \sum_{mTP \in \mathbb{TP}} \left( 1 - \min(1,mTP) \right) \right)
\label{eq:nds}
\end{equation}
The score is composed of two equal-weight halves. One half is the mean average precision ($mAP$), calculated over different object classes and matching distance thresholds. 

{\FH{
$mAP$ reflects the precision-recall trade-off across detection thresholds and object categories.
}}
The other half measures detection quality through bounding-box and attribute error metrics.
This is calculated through five types of error: translation, scale, orientation, velocity, and an other-attributes error term (e.g., whether a pedestrian is sitting or standing). 
$\mathbb{TP}$ denotes the set of mean errors $mTP$ for each of the five types. The errors are capped to 1 and the result is subtracted from 1, so that the score is maximized when the errors are all zero.

%
%
\subsection{Lightweight Features Offloading: Hybrid Computing}
\label{SubSec:lightweight_features_sharing}

In hybrid computing, multi-view images captured around the vehicle are first processed onboard by the initial backbone layers, \FH{extract intermediate feature vectors}. 
These features are then clipped and compressed to their dimensionality and transmission size before being transmitted to the cloud.
In the cloud, the remaining backbone processing, view transformation, \ac{bev} encoding, and 3D object detection are completed. 
This division reduces the computational load on the vehicle, while resource intensive tasks are handled in the cloud. 
Algorithm~\ref{algo:perception_loop} defines the perception loop, which periodically runs the perception task by calling upon the local and cloud routines.
\begin{algorithm}[ht]
\caption{Perception loop}
\label{algo:perception_loop}
\centering
\resizebox{0.95\linewidth}{!}{%
\begin{minipage}{\linewidth}
\begin{algorithmic}
\Require \\$nets$: set of $n$-layer backbone networks, one per precision/quantization level $q$ \\
         $split$: \FH{split layer index}, integer $\in \{1,\dots,n\}$ \\
         $q$: quantization level \\
         \FH{$cliPcen=(cliPcen_{low},cliPcen_{up})$}: clipping percentiles \\
         $\Delta t$: time period between consecutive perception runs
\end{algorithmic}
\setcounter{ALG@line}{0}
\begin{algorithmic}[1]
\Procedure{percLoop}{$nets$, $split$, $q$, $cliPcen$, $\Delta t$}
  \While{vehicle is driving}
    \State $t_{start} \gets \text{current time}$
    \State \FH{$net \gets nets_q$} \Comment{\FH{backbone corresponding to $q$}}
    \State $fVec_{comp} \gets \text{percLoc}(net, split, cliPcen)$
    \State $\text{percCloud}(fVec_{comp}, split, q)$ \Comment{\FH{cloud routine}}
    \State $t_{end} \gets \text{current time}$
    \Sleep{$\max\left(0,\Delta t - (t_{end}-t_{start})\right)$}
  \EndWhile
\EndProcedure
\end{algorithmic}
\end{minipage}
}
\end{algorithm}
\begin{algorithm}[ht]
\caption{On-board perception function}
\label{algo:onboard_perception_function}
\centering
\resizebox{0.95\linewidth}{!}{%
\begin{minipage}{\linewidth}
\begin{algorithmic}
 \Require  \\$net$: $n$-layer backbone network \FH{(configured at precision $q$)} \\
          $split$: \FH{split layer index}, integer $\in \{1,\dots,n\}$ \\
          \FH{$cliPcen=(cliPcen_{low},cliPcen_{up})$}: clipping percentiles
\Ensure $fVec_{comp}$: clipped and compressed feature vector
\end{algorithmic}
\setcounter{ALG@line}{0}
\begin{algorithmic}[1]
\Function{percLoc}{net, $split$, $cliPcen$}
  \State $imgData \gets \text{raw multi-view image input}$
  \State $fVec \gets imgData$
  \For{$l = 1$ to $split$} \Comment{execute onboard backbone layers}
    \State $fVec \gets net_{l}(fVec)$
  \EndFor
  \State $thres_{low} \gets \text{percentile}(fVec, cliPcen_{low})$
  \State $thres_{up} \gets \text{percentile}(fVec, cliPcen_{up})$
  \State $fVec_{clip} \gets \text{clip}(fVec, thres_{low}, thres_{up})$
  \State \FH{$fVec_{comp} \gets \text{compress}(fVec_{clip})$} \Comment{\FH{lossless (zlib)}}
  \State \Return $fVec_{comp}$
\EndFunction
\end{algorithmic}
\end{minipage}
}
\end{algorithm}

%
\textbf{On-board Component:}
Algorithm~\ref{algo:onboard_perception_function} defines the processing performed on the vehicle. 
It begins by acquiring the \FH{raw multi-view image data}. 
The backbone network \FH{$net^{(q)}$, configured at precision level $q$,} extracts an \FH{intermediate feature tensor} 
$fVec \in \mathbb{R}^{C \times H \times W}$, 
where $C$ \FH{denotes} the number of channels and $H$, $W$ represent the \FH{spatial height and width dimensions}. Computation proceeds up to the layer specified by the $split$ parameter. Layers beyond this point are executed in the cloud. The split point determines the balance between onboard and cloud computation.
Because the feature tensor typically decreases in spatial resolution and channel dimensionality across layers, the split depth also directly influences the amount of data transmitted to the cloud.
To \FH{reduce transmission overhead}, $fVec$ is \FH{percentile-clipped using thresholds} $cliPcen_{low}$ and $cliPcen_{up}$, where values outside this range are \FH{truncated}. 
Based on \FH{empirical evaluation}, the 10\textsuperscript{th} and 90\textsuperscript{th} percentiles \FH{provide a favorable trade-off between feature size reduction and detection accuracy}. 
\FH{Clipping reduces the dynamic range and entropy of the feature distribution, thereby improving compressibility.} The clipped \FH{feature tensor} is subsequently compressed using the lossless algorithm~\cite{zlib} before being transmitted to the cloud via C-V2X.

\begin{algorithm}[!t]
\caption{Cloud perception procedure}
\label{algo:cloud_perception_procedure}
\centering
\resizebox{0.95\linewidth}{!}{%
\begin{minipage}{\linewidth}
\begin{algorithmic}
\Require \\$nets$: set of $n$-layer backbone networks, one per \FH{precision/quantization level $q$} \\
         $fVec_{comp}$: compressed perception \FH{feature tensor} \\
         $split$: \FH{split layer index}, integer $\in \{1,\dots,n\}$ \\
         $q$: quantization level
\end{algorithmic}
\setcounter{ALG@line}{0}
\begin{algorithmic}[1]
\Procedure{percCloud}{\FH{$nets$,} $fVec_{comp}$, $split$, $q$}
  \State $net \gets nets_q$ \Comment{\FH{backbone corresponding to $q$}}
  \State $fVec \gets \text{decompress}(fVec_{comp})$
  \For{\FH{$l = split+1$ \textbf{to} $n$}} \Comment{\FH{execute remaining layers}}
    \State $fVec \gets net_{l}(fVec)$
  \EndFor
  \State $\text{percepRes} \gets \text{BEVFormer.remainder}(fVec)$
  \State $\text{CPM} \gets \text{encode}(\text{percepRes})$
  \State \textbf{broadcast} $\text{CPM}$
\EndProcedure
\end{algorithmic}
\end{minipage}
}
\end{algorithm}

\textbf{Cloud Component:} 
Algorithm~\ref{algo:cloud_perception_procedure} specifies the perception operations \FH{executed} in the cloud.
The procedure begins by decompressing the received feature tensor and instantiating the backbone corresponding to precision level $q$.
The remaining backbone layers (i.e., layers $split+1$ to $n$) are then executed to complete feature extraction.
A multi-GPU setup (see Section~\ref{subSec:test_route}) is employed for this \FH{stage}, \FH{significantly accelerating processing compared to onboard execution}. 
Upon completion of the backbone, the \FH{remaining BEVFormer modules (view transformation, BEV encoding, and detection head)} are executed, \FH{resulting in a set of 3D bounding boxes}.
Finally, \FH{as in the onboard-only scenario}, the detection results are encoded into a \ac{cpm} and transmitted via C-V2X to nearby vehicles and infrastructure, \FH{thereby enabling cooperative perception}.
%
\subsection{CPM Encoding}
\label{subSec:methodology_cpm_encoding}
\begin{figure}[!t]
    \centering
    \includegraphics[width=0.99\linewidth]{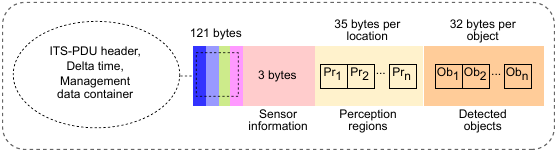}
    \caption{A basic overview of different containers included in the CPM message format as defined by the ETSI standard \cite{ts2023103}.}
    \label{fig:cpm_format}
\end{figure}

The \ac{cpm} encoding process packages detected objects and environmental data into a standardized format defined by ETSI \cite{ts2023103}, ensuring \ac{cps} interoperability. 
As illustrated in Figure~\ref{fig:cpm_format}, the message structure includes several containers such as the ITS-PDU header, management, and sensor information containers, which store the reference position, sensor ID, and metadata.
The use of ETSI-compliant CPM messages ensures compatibility with existing cooperative perception systems and facilitates integration into practical V2X deployments.
The encoding process is managed by the YoGoKo Y-Box module \cite{robocar}, which supports both ITS-G5 and C-V2X technologies, enabling seamless V2X communication. 

\section{Experiments and Results}
\label{sec:experiments_results}
In this section we assess the performance of our proposed 3D perception system, using the setup described in Section~\ref{sec:methodology}. Each experiment was repeated five times, for statistical robustness.  

\subsection{Onboard Computing and CPM Transmission}
\label{subSection:onboard_processing_cpm_transmission}
To establish a baseline, we performed inference tests on the onboard platform, as detailed in Section~\ref{subSec:hardware_configuration}, using the nuScenes dataset described in~\ref{SubSec:dataset_evaluation}. 
These tests yielded an average inference time of \ms{673} for the default model prior to optimization.
This far exceeds the typical latency threshold for real-time perception in autonomous driving of \ms{100}~\cite{hawlader2024leveraging}. 
The \ms{100} bound is widely adopted in the literature as a practical upper limit for maintaining safe and responsive perception-driven control in urban driving scenarios.
Although the model achieved an \ac{nds} of 0.52, onboard processing consumed over \pcent{65} (±4) of the hardware resources, as monitored through the \texttt{nvidia-smi} GPU tracking tool. 
\FH{Such sustained utilization levels indicate limited headroom for additional perception or control tasks, further constraining real-time operation.}
These results highlight the limitations of onboard computing, particularly regarding resource usage and latency.

\textbf{Model Optimization using TensorRT:}
To address resource usage and latency, we optimized the model with TensorRT~\cite{10483824}. 
TensorRT enhances performance by applying precision calibration, or quantization, (FP32, FP16, or FP8, with the numeric suffix indicating how many bits are used) and reducing the model's computational complexity.
The experimental results in Table~\ref{table:tensorrt_optimization} show how TensorRT optimization significantly decreased inference times, without major degradation in detection accuracy. 
For instance, inference time dropped from \ms{486} with FP32 to \ms{194} with FP8, with only a marginal decrease in \ac{nds} from 0.52 to 0.51, indicating that detection performance was largely preserved even with reduced precision. 
These results align with previous studies, which demonstrated that quantization effectively maintains high detection accuracy while significantly reducing inference time \cite{10184097,10001050}.
While quantization substantially improves onboard performance, the resulting latency still remains above stringent real-time perception targets, motivating the need for hybrid offloading.
\begin{table}[!h]
\centering
\resizebox{0.92\linewidth}{!}{%
\begin{tabular}{c c c c c}
\textbf{Quantization} & \textbf{Inference (ms)} & \textbf{NDS} & \textbf{CPM (ms)} & \textbf{\shortstack{End-to-end \\ delay (ms)}} \\ 
\midrule
\rowcolor{gray!10} FP32 & 486 & 0.52 & 5.9 $(\pm 1.8)$ & 491.9 $(\pm 2.7)$\\ 
FP16 & 257 & 0.52 & 5.7 $(\pm 1.7)$ & 262.7 $(\pm 2.3)$ \\ 
\rowcolor{gray!10} FP8 & 194 & 0.51 & 4.9 $(\pm 1.6)$ & 198.9 $(\pm 2.3)$ \\ 
\bottomrule
\end{tabular}
}
\caption{Performance metrics for the BEVFormer model with ResNet101 backbone, evaluated using TensorRT optimization at different quantization levels (FP32, FP16, FP8) on the onboard vehicle platform, as detailed in Section~\ref{subSec:hardware_configuration}. The table includes inference time and CPM transmission latency, which together form the end-to-end delay. Standard deviations $(\pm)$ are provided to reflect measurement variability.}
\label{table:tensorrt_optimization}
\end{table}

\textbf{CPM Transmission \& V2X Communication:} Upon detecting surrounding objects, the 3D detection results are encoded into a \ac{cpm} and broadcast to nearby vehicles via ITS-G5. 
We performed a CPM transmission test to measure the associated end-to-end latency in a cooperative perception scenario. 
Table \ref{Tab:network_parameters} summarizes the network configuration used. 
\FH{The experiments were conducted under real vehicular mobility conditions to capture practical channel variability and interference characteristics of commercial ITS-G5 deployments.}
\begin{table}[!h]
\centering 
\resizebox{.85\linewidth}{!}{%
\begin{tabular}{ll}
\textbf{Parameter Name} & \textbf{Value}  \\ 
\midrule
Transmission Power (Tx) & 23 dBm \\
Energy threshold & -85 dBm \\
Channel bandwidth / carrier frequency & \MHz{10} / \GHz{5.9} \\
Radio Configuration & Single Channel (CCH) \\
Data rate & \Mbitps{7} \\
Number of CPM Transmitted / loss ratio & 6000 / 0.09 \\
\midrule
\end{tabular}}
\caption{Important network parameters for V2X.}
    \label{Tab:network_parameters}
\end{table}

\begin{figure}[t]
    \centering
    \includegraphics[width=\linewidth, trim={0cm 0.25cm 0 0.5cm}, clip]{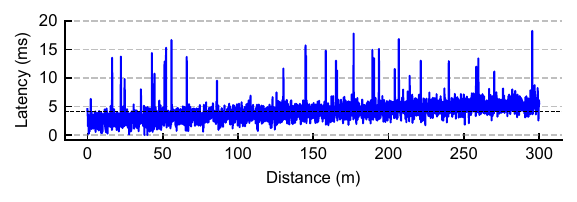}
    \caption{CPM transmission latency versus distance between a moving vehicle (\kmh{25} avg.) and a stationary receiver at fixed coordinates (longitude: 6.161993, latitude: 49.626478). Dashed line shows average latency \ms{4.10}.}
    \label{fig:cpm_tranmission_latency_v2v_communication}
\end{figure}

A single static receiver node was placed at a fixed location, stated in the caption of Figure~\ref{fig:cpm_tranmission_latency_v2v_communication}. 
The sender node was placed in a moving vehicle, as detailed in Section \ref{subSec:test_route}.

The results, shown in Figure~\ref{fig:cpm_tranmission_latency_v2v_communication}, illustrate how \acp{cpm} transmission latency varies with the distance between the two communicating nodes.
The results indicate a slight, linear increase in transmission latency as distance grows, likely due to propagation delay and intermittent network congestion. The average transmission latency was \ms{4.10}, with a maximum of \ms{18.41} and a standard deviation of \ms{1.61}.
Notably, packet loss increased significantly when distance exceeded \m{300}, highlighting sensitivity to longer distances. These observations are consistent with previous simulation-based research \cite{10184097}.

\begin{figure}[!t]
    \centering
    \includegraphics[width=\linewidth]{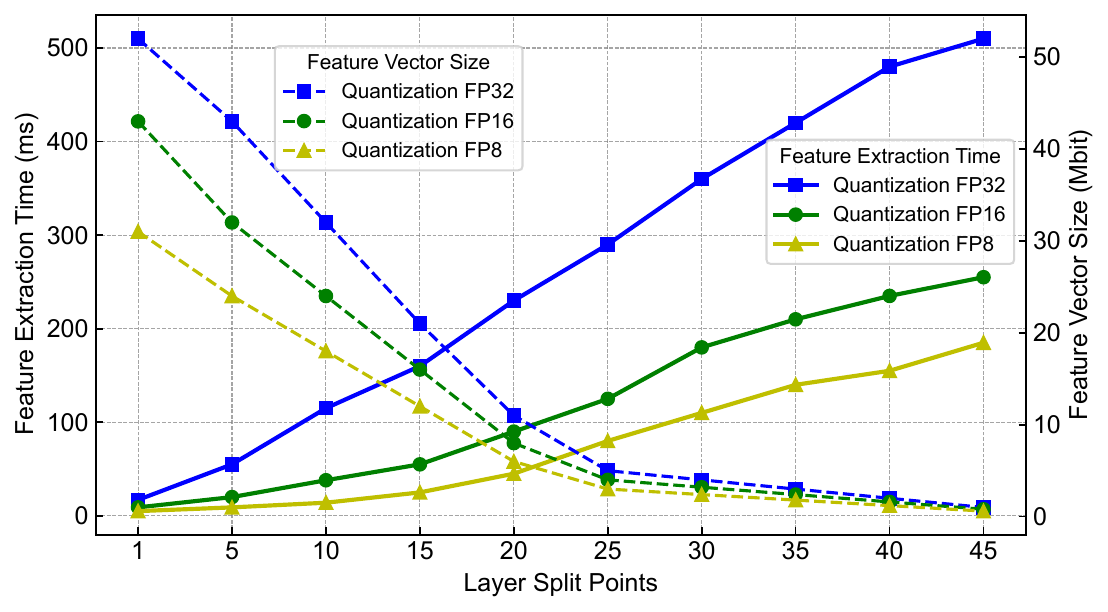}
    \caption{Feature vector size and extraction time versus split depth. Solid lines represent feature extraction time (left y-axis), while dashed lines indicate feature size (right y-axis). Lower-precision quantization (e.g., FP16, FP8) reduces both extraction time and feature size.}
    \label{fig:latency_feature_vector_plot}
\end{figure}

Although TensorRT helped significantly reduce end-to-end delay, as shown in Table~\ref{table:tensorrt_optimization}, onboard processing still falls short of the \ms{100} real-time perception target.
For example, FP8 quantization yields an end-to-end delay (onboard inference plus transmission) of \ms{198.9}, which is almost double the desired threshold. 
Such latency corresponds to an effective perception update rate of approximately 5~Hz, significantly below the 10~Hz rate typically associated with \ms{100} class perception systems.
Consequently, the end-to-end delay restricts \ac{cpm} transmission rates to below \Hz{5}, highlighting the need for a more efficient solution to support real-time perception.
\subsection{Hybrid Computing and Lightweight Features Sharing}
\label{subSection:cloud_processing_results}
Offloading intensive perception tasks to the cloud, while keeping lighter tasks onboard, reduces local computation but adds transmission latency. 
Techniques like post-training quantization, compression, and clipping help minimize bandwidth usage and transmission time. 
This section explores the feasibility of lightweight feature sharing over the network, focusing on split layer selection, accuracy retention, and end-to-end delay.

\textbf{Layer Partitioning and Feature Extraction:}
In the hybrid computing case, determining the optimal backbone partition layer is crucial, as it affects both the onboard feature extraction time and the size of transmitted features. 

In BEVFormer, partitioning earlier, e.g., layer 1, minimizes onboard computation but requires transmitting larger feature vectors to the cloud. 
{\FH{Conversely, partitioning later reduces the size of the transmitted feature vector $S_{feat}(split,q)$---defined as the compressed payload (in bits) generated at the selected split point and quantization level---at the cost of additional onboard inference time.}}
Figure~\ref{fig:latency_feature_vector_plot} illustrates the trade-off between inference time and feature vector size for different split points and quantization levels.
\begin{itemize}
    \item \textbf{Feature extraction time:} As more backbone layers are executed locally, feature extraction time increases, significantly impacting real-time perception feasibility. A split at layer 5 with FP32 quantization yields a latency of \ms{55}, acceptable for real-time use, but deeper splits quickly exceed the \ms{100} threshold (e.g., \ms{115} at layer 10). Lower precisions like FP16 (\ms{20} at layer 5) and FP8 (\ms{9} at layer 5) reduce latency, but the benefits diminish with deeper splits, where even FP8 exceeds \ms{140} by layer 30. 
    These results indicate that split depth has a dominant influence on onboard latency, and aggressive quantization alone cannot compensate for excessive local computation.
    \item \textbf{Feature vector size:} Shallow or earlier splits generate large feature vectors, making data transmission slow. For example, FP32 split after layer 1 produces \MB{52}, requiring a throughput of \Mbitps{520} at \Hz{10}, far exceeding typical V2X bandwidth. Even FP16 (\MB{43}, \Mbitps{430}) and FP8 (\MB{31}, \Mbitps{310}) require too much bandwidth. Deeper splits, however, reduce vector sizes: FP32 split after layer 25 requires \MB{5} (\Mbitps{50}), and FP8 only \MB{3} (\Mbitps{30}), which are more suitable for real-time transmission. Nevertheless, deeper splits increase onboard processing time, creating a fundamental latency bandwidth trade-off.
    This trade-off highlights the need for adaptive split selection under dynamic bandwidth constraints, motivating the dynamic parameter optimization scheme introduced in the next Section~\ref{sec:dynamic_offloading_parameter_selection}.
\end{itemize}
\begin{figure}[!t]
    \centering
    \includegraphics[width=\linewidth]{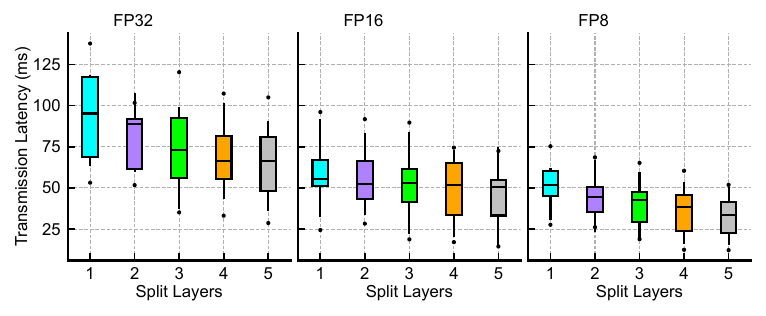}
    \caption{Transmission latency of feature vectors from vehicle to cloud across five split layers for FP32, FP16, and FP8 over a 5G network using C-V2X. FP8 demonstrates the lowest and most stable latency, suitable for real-time transmission. FP32 exhibits the highest latency and variability, especially at earlier split layers due to larger feature size.}
    \label{fig:latency_box_plot}
\end{figure}
\begin{table*}[!t]
\centering
\resizebox{0.95\textwidth}{!}{%
\begin{tabular}{cc|cc|cc|cc|ccc}
\multirow{2}{*}{\textbf{FP}} & \multirow{2}{*}{\textbf{\shortstack{Split \\ layer}}} & \multicolumn{2}{c|}{\textbf{Onboard Processing Time (ms)}} & \multicolumn{2}{c|}{\textbf{Transmission latency (ms)}} & \multicolumn{2}{c|}{\textbf{Cloud Processing Time (ms)}} & \multirow{2}{*}{\textbf{\shortstack{End-to-end \\ Delay (ms)}}} & \multirow{2}{*}{\textbf{NDS}} & \multirow{2}{*}{\textbf{\shortstack{Bandwidth \\ Usage (\Mbitps{})}}} \\ 
                  &                       & \textbf{Backbone} & \textbf{Compression} & \textbf{V2C} & \textbf{C2V} & \textbf{Decompression} & \textbf{Head} &  &  &  \\ \hline

\multirow{5}{*}{32} 
& 1  & 17.2 $(\pm 2.10)$ & 10.7 $(\pm 1.85)$   & 65.8 $(\pm 4.00)$  & 11.6  $(\pm 1.15)$   & 2.6 $(\pm 0.60)$  & 20.8  $(\pm 1.35)$   & 128.7 $(\pm 4.20)$  & 0.52 & 10.5 \\ 
& 2  & 22.3 $(\pm 1.75)$ & 8.6  $(\pm 1.55)$   & 58.0 $(\pm 3.90)$  & 9.8   $(\pm 0.95)$   & 2.4 $(\pm 0.55)$  & 18.4  $(\pm 1.25)$   & 119.6 $(\pm 3.90)$  & 0.50 & 8.4 \\ 
& 3  & 30.5 $(\pm 1.90)$ & 7.3  $(\pm 1.50)$   & 48.9 $(\pm 3.75)$  & 8.4   $(\pm 0.85)$   & 2.5 $(\pm 0.50)$  & 15.9  $(\pm 1.30)$   & 113.5 $(\pm 3.80)$  & 0.48 & 6.8 \\ 
& 4  & 39.8 $(\pm 1.65)$ & 6.4  $(\pm 1.30)$   & 54.5 $(\pm 3.50)$  & 7.0   $(\pm 0.75)$   & 2.3 $(\pm 0.45)$  & 14.7  $(\pm 1.10)$   & 124.7 $(\pm 3.60)$  & 0.47 & 5.9 \\ 
& 5  & 55.4 $(\pm 1.45)$ & 5.1  $(\pm 1.10)$   & 56.3 $(\pm 3.20)$  & 5.8   $(\pm 0.70)$   & 2.5 $(\pm 0.40)$  & 12.6  $(\pm 0.95)$   & 137.7 $(\pm 3.40)$  & 0.46 & 5.4 \\ 
\hline
\multirow{5}{*}{16} 
& 1  & 9.3  $(\pm 1.70)$ & 9.1  $(\pm 1.50)$ &57.6 $(\pm 3.10)$ & 12.7 $(\pm 0.95)$ & $(\pm 0.50)$ & 18.2 $(\pm 1.30)$ & 109.0 $(\pm 3.90)$ & 0.51 & 9.0 \\
& \gcell{2}  & \gcell{11.7 $(\pm 1.50)$} & \gcell{7.3  $(\pm 1.30)$} & \gcell{39.3 $(\pm 2.95)$} & \gcell{7.6  $(\pm 0.85)$} & \gcell{2.2 $(\pm 0.45)$} & \gcell{16.6 $(\pm 1.20)$} & \gcell{84.7  $(\pm 3.70)$} & \gcell{0.49} & \gcell{6.6} \\
& \gcell{3}  & \gcell{15.3 $(\pm 1.35)$} & \gcell{6.2  $(\pm 1.20)$} & \gcell{44.1 $(\pm 2.80)$} & \gcell{6.6  $(\pm 0.80)$} & \gcell{2.1 $(\pm 0.40)$} & \gcell{14.3 $(\pm 1.05)$} & \gcell{88.7  $(\pm 3.50)$} & \gcell{0.47} & \gcell{5.6} \\
& \gcell{4}  & \gcell{18.5 $(\pm 1.25)$} & \gcell{5.2  $(\pm 1.05)$} & \gcell{42.3 $(\pm 2.65)$} & \gcell{8.6  $(\pm 0.75)$} & \gcell{2.0 $(\pm 0.35)$} & \gcell{13.4 $(\pm 0.95)$} & \gcell{90.0  $(\pm 3.25)$} & \gcell{0.46} & \gcell{4.6} \\
& \gcell{5}  & \gcell{20.4 $(\pm 1.10)$} & \gcell{4.3  $(\pm 0.95)$} & \gcell{31.2 $(\pm 2.50)$} & \gcell{7.1  $(\pm 0.70)$} & \gcell{2.0 $(\pm 0.30)$} & \gcell{12.2 $(\pm 0.85)$} & \gcell{77.2  $(\pm 3.05)$} & \gcell{0.45} & \gcell{4.3} \\ 
\hline

\multirow{5}{*}{8} 
& \gcell{1}  & \gcell{5.1  $(\pm 1.45)$} & \gcell{8.2  $(\pm 1.45)$} & \gcell{33.6 $(\pm 2.80)$} & \gcell{9.8  $(\pm 0.90)$} & \gcell{1.6 $(\pm 0.50)$} & \gcell{15.5 $(\pm 1.25)$} & \gcell{73.8  $(\pm 3.90)$} & \gcell{0.47} & \gcell{8.4} \\
& \gcell{2}  & \gcell{6.2  $(\pm 1.25)$} & \gcell{6.7  $(\pm 1.30)$} & \gcell{40.4 $(\pm 2.60)$} & \gcell{8.1  $(\pm 0.85)$} & \gcell{1.6 $(\pm 0.45)$} & \gcell{14.1 $(\pm 1.15)$} & \gcell{77.1  $(\pm 3.60)$} & \gcell{0.46} & \gcell{6.9} \\
& \gcell{3}  & \gcell{7.3  $(\pm 1.10)$} & \gcell{5.7  $(\pm 1.10)$} & \gcell{44.3 $(\pm 2.40)$} & \gcell{6.3  $(\pm 0.80)$} & \gcell{1.5 $(\pm 0.40)$} & \gcell{12.6 $(\pm 1.00)$} & \gcell{77.7  $(\pm 3.50)$} & \gcell{0.44} & \gcell{5.5} \\
& \gcell{4}  & \gcell{8.4  $(\pm 1.05)$} & \gcell{4.7  $(\pm 1.00)$} & \gcell{33.4 $(\pm 2.20)$} & \gcell{5.6  $(\pm 0.75)$} & \gcell{1.5 $(\pm 0.35)$} & \gcell{11.9 $(\pm 0.90)$} & \gcell{65.5  $(\pm 3.25)$} & \gcell{0.43} & \gcell{4.7} \\
& \gcell{5}  & \gcell{9.1  $(\pm 0.90)$} & \gcell{3.6  $(\pm 0.90)$} & \gcell{29.3 $(\pm 2.00)$} & \gcell{7.0  $(\pm 0.70)$} & \gcell{1.4 $(\pm 0.30)$} & \gcell{11.0 $(\pm 0.85)$} & \gcell{61.9  $(\pm 3.00)$} & \gcell{0.43} & \gcell{4.1} \\
\hline
\end{tabular}%
}
\caption{Performance metrics for various split layers and quantization levels, including onboard processing time, V2C and C2V transmission latency, cloud processing time, and total end-to-end delay. The standard deviations $(\pm)$ reflect variability. The last column shows bandwidth utilization for offloading feature vectors from vehicle to cloud. Rows with end-to-end delay below 100~ms are highlighted in green.}
\label{tab:delay_performance_breakdown}
\end{table*}

\textbf{Feature Compression and Transmission:}
To reduce feature vector sizes beyond what quantization can offer, we employed dynamic clipping and zlib compression.
These techniques reduced feature size by approximately \pcent{97} for FP32, \pcent{90} for FP16, and \pcent{80} for FP8, significantly improving transmission efficiency across the board.
The positive correlation between encoding bits and compressibility means that, in this dataset, entropy grows sublinearly with the number of bits used, allowing the compression algorithm to take advantage of the added repetition or structure in the data.
Figure~\ref{fig:latency_box_plot} shows the observed transmission latency for different quantization and split layer combinations. 
We focus on layers 1 to 5 because deeper splits yielded minimal improvements in transmission latency, while increasing feature extraction time.
FP8 exhibited the lowest latency, with medians of \ms{52} at layer 1 and \ms{35} at layer 5, meeting the threshold required for real-time perception systems.
In contrast, FP32 exhibited the highest latency, reaching \ms{90} at layer 1 and \ms{70} at layer 5. Variance was also the highest among the different quantization levels. 
FP16 provided a good middle ground, with median latencies of \ms{67} at layer 1 and \ms{45} at layer 5, making it suitable for applications that can tolerate moderate delays.
FP8, with its consistently low latency, is ideal for latency-critical applications communicating over low bandwidth-networks, whereas FP16 and FP32 are more suitable for faster networks, or scenarios with more relaxed latency bounds.
\FH{These observations confirm that quantization level and split depth jointly determine the feasible latency region, reinforcing the need for adaptive configuration under dynamic bandwidth conditions.}

\textbf{End-to-End Delay:}  
The results in Table \ref{tab:delay_performance_breakdown} detail the impact of split layer and quantization level choice on the different latency components.  
For FP32 quantization at split layer 1, the total end-to-end delay was \ms{128.7}, with local processing time (backbone and compression) contributing \ms{27.9}, and transmission latency (V2C and C2V) adding \ms{77.4}.  
In contrast, FP8 at the same split layer yielded a significantly lower total delay of \ms{73.8}, mainly due to the shorter local processing time of \ms{13.3} and a reduced transmission latency of \ms{43.4}.  
Lower quantization levels, such as FP8, reduce both computational burden and transmission latency, though they introduce a slight trade-off in accuracy, with the \ac{nds} for FP8 at split layer 1 being 0.47, compared to 0.52 for FP32.  

As the split layer deepens (e.g., layer 5), onboard processing time increased, due to the more complex feature extraction.  
For FP32, the total delay at split layer 5 grew to \ms{137.7}, with \ms{60.5} for local processing and \ms{62.1} for data transmission.  
In comparison, FP8 achieved a lower end-to-end delay of \ms{61.9} at the same split layer, primarily due to its shorter local processing time of \ms{12.7} and transmission latency of \ms{36.3}.  
FP8 at split layer 5 achieved the lowest end-to-end delay (\ms{61.9}) and bandwidth usage (\Mbitps{4.1}), with only a marginal drop in accuracy (\ac{nds} = 0.43).
Cloud processing times (decompression and detection head) remained consistently low across all quantization levels due to the powerful hardware, with decompression ranging from 1.4 to \ms{2.6}. 
This stability ensured that the majority of delay originated from local processing and transmission, emphasizing the need to carefully select the optimal split point and quantization level for real-time systems. 
Transmission latencies showed greater variability than local or cloud processing, with the highest deviations observed at split layer 1 for FP32 ($\pm$\ms{4.00} V2C) and FP8 ($\pm$\ms{2.80} V2C), reflecting the influence of fluctuating network conditions. 
While lower precisions like FP8 reduced total delay, they introduced minor accuracy degradation a trade-off further examined in the following section.

\begin{figure}[!t]
    \centering
    \includegraphics[width=\linewidth]{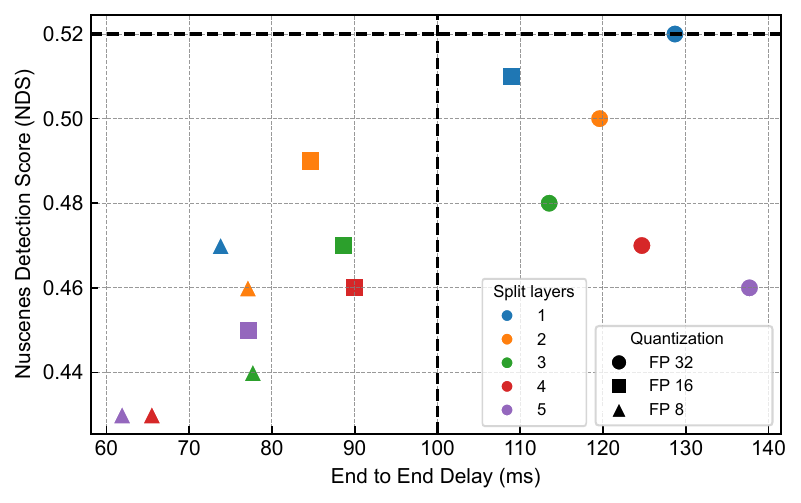}
    \caption{End-to-end delay vs. detection accuracy (\ac{nds}) across split layers and quantization levels. Earlier layers result in higher accuracy but increased delay, while intermediate layers provide a balance between delay and accuracy.}
    \label{fig:nsd_end2end_latency}
\end{figure}

\begin{figure*}[!t]
\centering
\includegraphics[width=\linewidth]{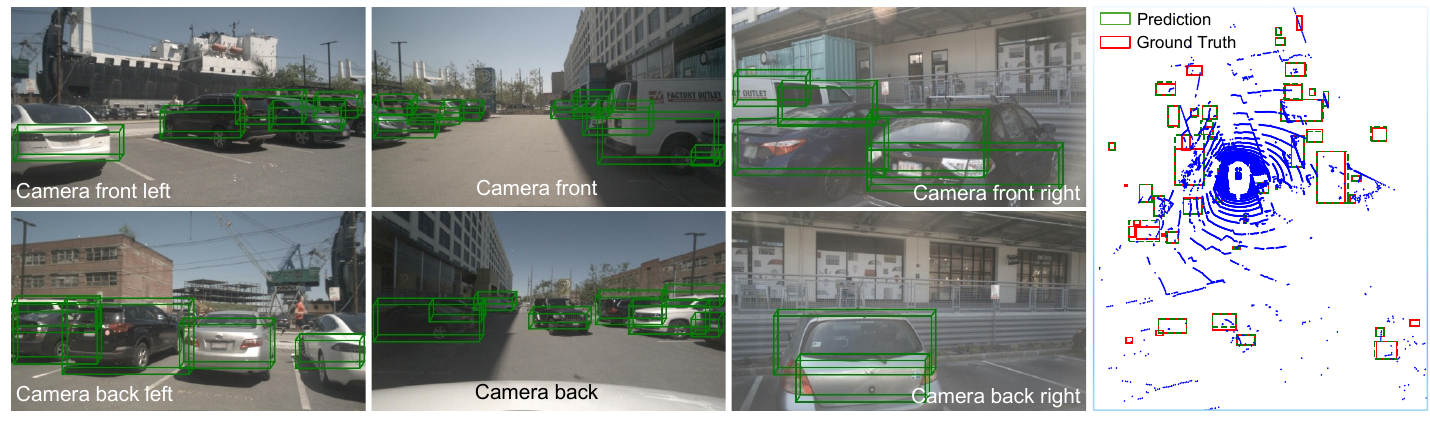}
\caption{Qualitative results at split layer 3 with FP16 on the nuScenes validation set (SceneID: n008-2018-05-21-11-06-59-0400), an \ac{nds} score of 0.47. We show 3D bounding box predictions in multi-view (6x) camera images and the 360-degree \ac{bev}.}
\label{fig:visualization}
\end{figure*}

\textbf{End-to-End Delay vs. Accuracy Trade-off:}  
Figure \ref{fig:nsd_end2end_latency} illustrates the trade-off between end-to-end delay and \ac{nds} across different split layers and quantization levels.  
A clear trend is that deeper split points lead to reduced detection accuracy, regardless of the selected quantization level. 
This suggests that clipping and quantization applied to deeper feature representations may remove information that has become more semantically compact and task-relevant.
As network depth increases, feature representations become progressively more compressed and discriminative, so additional information loss due to reduced precision or clipping has a proportionally larger impact on detection performance.

The other major trend is that increasing the quantization level, while holding the split layer constant, increased both delay and detection accuracy alike. From this it follows that (split=1, FP32) yielded the highest accuracy and delay, and (split=5, FP8) the lowest accuracy, delay combination.
A particularly well-performing combination was (split=1, FP16), which yielded \pcent{98} of (split=1, FP32)'s accuracy, but only \pcent{85} of the delay.
The (split=3, FP16) combination, a visual demonstration of which is shown in Figure~\ref{fig:visualization}, offered an \ac{nds} of 0.47 and a delay of \ms{88.7}. 
This configuration satisfies commonly adopted sub-\ms{100} real-time perception bounds while preserving competitive detection accuracy.
This makes it a practical solution for perception systems with a \ms{100} latency bound.

As shown in Table~\ref{tab:delay_performance_breakdown}, the impact of changing the split layer on end-to-end delay is not straightforward, even when the quantization level is held constant. 
For instance, with FP32 quantization, splitting at layer 2 results in a lower delay (119.6 ms) compared to layer 1 (128.7 ms), despite the deeper layer. 
Conversely, moving from layer 3 to layer 4 increases the delay from 113.5 ms to 124.7 ms, highlighting the non-monotonic nature of the trade-off. 
This behavior stems from a complex interplay between feature vector size, onboard computation, and transmission time. 
Consequently, static configurations are often suboptimal under real-world network conditions that exhibit fluctuating bandwidth and latency. 
\FH{To address this limitation, we introduce a dynamic parameter selection mechanism that jointly adapts the split layer and quantization level based on real-time bandwidth availability.}
The objective is to maximize detection accuracy while satisfying a strict end-to-end latency budget, as detailed in Section~\ref{sec:dynamic_offloading_parameter_selection}.
\section{Dynamic Hybrid-Computing Parameter Selection}
\label{sec:dynamic_offloading_parameter_selection}

We have shown that varying the split point and quantization level leads to different trade-offs among bandwidth usage, end-to-end detection latency, and perception accuracy. 
In this section, we propose a dynamic selection algorithm that jointly optimizes these two parameters to maximize detection performance while adhering to bandwidth and latency constraints.

\subsection{Problem Formulation}

The goal is to select the split point $split$ and quantization $q$ parameters that maximize the \ac{nds} accuracy metric from Eq.~\ref{eq:nds}:
\begin{equation}
(split, q) = \arg\max_{split, q} NDS(split,q).
\end{equation}

This maximization must be subject to two constraints:

\textbf{Bandwidth budget (\mbox{\boldmath$bw_{upl}$}, \mbox{\boldmath$bw_{dwn}$}):} the available uplink and downlink bandwidth for the offloading process. This may reflect total channel capacity or a reserved slice allocated for the perception task, depending on network sharing policies, and leaving room for other applications. 
 
\textbf{Latency limit (\mbox{\boldmath$lat_{max}$}):} the total amount of time available for the perception task. Ideally set to \ms{100} for real-time cooperative perception~\cite{hawlader2024leveraging}.
The total perception latency, $lat_{total}$, which is a function of the split point, the quantization level, and the bandwidth budget, must therefore not be larger than $lat_{max}$:
\begin{equation}
 lat_{total}(split,q,bw_{upl},bw_{dwn}) \leq lat_{max}.
\label{eq:latency_bound}
\end{equation}
This constraint, together with the bandwidth budget, forms the optimization boundary.
The total latency can be decomposed into the sum of four distinct phases:
\begin{equation}
\begin{split}
lat_{total}(split,q,bw_{upl},bw_{dwn}) & = lat_{local}(split,q) \\
& + lat_{upl}(split, q, bw_{upl}) \\
& + lat_{cloud}(split, q) \\
& + lat_{dwn}(split, q, bw_{dwn}).
\end{split}
\label{eq:latency_total}
\end{equation}

$lat_{local}$ is the latency incurred locally, computing the BEVFormer backbone layers pre-split, clipping, and compressing the feature vector. 
$lat_{upl}$ \FH{and $lat_{dwn}$ denote the uplink and downlink transmission times, respectively. 
These terms are modeled using an effective transfer time that is measured or estimated online. 
This formulation captures the dominant delay components in terrestrial V2X and cellular systems, including radio access scheduling, medium access contention, retransmissions, and queuing effects under vehicular mobility. 
Propagation delay is negligible in terrestrial deployments~\cite{sullivan2012overview}.} 

\FH{Even over tens of kilometers, propagation delay remains sub-millisecond, on the order of a few microseconds per kilometer in fiber or radio links.
Consequently, propagation delay is excluded from the explicit decomposition in Eq.~\ref{eq:latency_total}. 
Mobility induced variability in end-to-end communication delay is implicitly captured in the effective transfer time.
As a result, the optimization remains valid under dynamic vehicular conditions and does not compromise real-world applicability.}
$lat_{cloud}$ is the time required to complete the remaining perception tasks in the cloud.
All latency components depend on the selected split point and quantization level, while the communication terms additionally depend on the available bandwidth budget.

\subsection{Dynamic Optimization Algorithm}
We begin by constructing an array \FH{$pArray$} that enumerates all valid $(split, q)$ parameter tuples. 
This array is pre-sorted in non-increasing order of \FH{\ac{nds}}, based on \FH{the empirical accuracy measurements} in Table~\ref{tab:delay_performance_breakdown}, such that higher-ranked configurations yield better detection accuracy.
During each iteration of the perception loop, the \FH{effective available uplink and downlink bandwidths} are estimated. 
The optimization process then selects the first (i.e., highest-ranked) tuple in $pArray$ that satisfies the latency constraint defined in Equation~\ref{eq:latency_bound}, where $lat_{max}$ corresponds to the perception cycle period $\Delta t$.
\FH{Given the profiled accuracy and latency components, this approach selects the most accurate configuration that meets the real-time latency constraint, thereby maximizing detection performance under dynamic network conditions.}

Algorithm~\ref{algo:perception_loop_adaptive_version} illustrates how this parameter optimization mechanism is embedded into the hybrid perception loop. 
Modifications relative to the baseline implementation (Algorithm~\ref{algo:perception_loop}) are underlined for clarity.
Specifically, the fixed $split$ and $q$ inputs are replaced with the adaptive parameter array $pArray$. 
In each loop iteration, the \FH{estimated uplink and downlink bandwidths}, denoted $bw_{upl}$ and $bw_{dwn}$, \FH{are used as inputs to the optimization}. 
While the bandwidth estimation algorithm itself is outside the scope of this paper, we assume the existence of a reliable estimator.
The current bandwidth estimates, along with $pArray$ and the latency constraint $\Delta t$, are passed to the function $optPar()$, which selects the \FH{split layer and quantization level} that satisfy the constraint. 

The selected parameters are then used in the subsequent perception operations, which otherwise remain unchanged.
Algorithm~\ref{algo:hybrid_computation_parameter_optimization} specifies the behavior of the parameter optimization function $optPar()$. 
The correctness of the optimization function is established by the following theorem, ensuring it selects parameters that maximize accuracy within the latency bound, or minimize delay if the bound is unattainable.

\begin{algorithm}[!b]
\caption{Perception loop (adaptive version)}
\label{algo:perception_loop_adaptive_version}
\centering
\resizebox{0.95\linewidth}{!}{%
\begin{minipage}{\linewidth}
\begin{algorithmic}
\Require \\$nets$: set of $n$-layer backbone networks, one per quantization \\
         \underline{$pArray$:} array of valid parameter tuples $(split, q)$, sorted in non-increasing \ac{nds} order\\
         $cliPcen$: clipping percentiles \\
         $\Delta t$: time period between consecutive perception runs
\end{algorithmic}
\setcounter{ALG@line}{0}
\begin{algorithmic}[1]
\Procedure{percLoop}{$nets$,\underline{$pArray$,}$cliPcen$, $\Delta t$}
  \While{vehicle is driving}
    \State $t_{start} \gets \text{current time}$
    \State \underline{$(bw_{upl}, bw_{dwn}) \gets \text{estimateBandwidth}()$}
    \State \underline{$(split, q) \gets optPar(pArray, bw_{upl}, bw_{dwn}, \Delta t)$}
    \State \FH{$net \gets nets_q$}
    \State \FH{$fVec_{comp} \gets \text{percLoc}(net, split, cliPcen)$}
    \State $\text{percCloud}(fVec_{comp}, split, q)$ \Comment{\FH{remote call}}
    \State $t_{end} \gets \text{current time}$
    \Sleep{$\max\left(0, \Delta t - (t_{end}-t_{start})\right)$}
  \EndWhile
\EndProcedure
\end{algorithmic}
\end{minipage}
}
\end{algorithm}

\begin{theorem}
$optPar()$ returns \FH{the} highest \ac{nds}-yielding parameter tuple $(split, q)$ that satisfies the latency bound $lat_{total} \leq lat_{max}$ or, if no such tuple exists, the tuple that minimizes $lat_{total}$.
\label{theo:optpar_correctness}
\end{theorem}

First, consider the case where at least one latency-bound-satisfying (i.e., acceptable) tuple exists. 
\texttt{optPar()} traverses the parameter tuple array sequentially from front to back. 
For each $(split, q)$ combination, it estimates the total latency $lat_{total}$ using Eq.~\ref{eq:latency_total}, with latency values derived from empirical measurements in Table~\ref{tab:delay_performance_breakdown}.
It then compares $lat_{total}$ with the maximum acceptable latency $lat_{max}$. 
The first tuple that satisfies the latency constraint is returned. 
Since the parameter array is sorted in non-increasing \ac{nds} order, this guarantees the returned tuple yields the highest \ac{nds} among all acceptable configurations, thereby proving the first part of Theorem~\ref{theo:optpar_correctness}.
Now consider the scenario where no tuple satisfies the latency constraint. 
To demonstrate the correctness of the algorithm in this case, we define the following loop invariant: 
at the start of each iteration \FH{with index $idx$}, $latMin$ holds the lowest latency observed among all tuples \FH{examined so far (i.e., indices $0$ to $idx-1$)}, and $latMinIdx$ stores the index of the corresponding parameter tuple.
We prove that this invariant holds by induction:

\begin{algorithm}[!t]
\caption{Hybrid-computing parameter optimization}
\label{algo:hybrid_computation_parameter_optimization}
\centering
\resizebox{0.95\linewidth}{!}{%
\begin{minipage}{\linewidth}
\begin{algorithmic}
 \Require \\$pArray$: non-empty \ac{nds}-sorted array of par. tuples $(split,q)$ \\
          $bw_{upl}$: upload bandwidth \\
          $bw_{dwn}$: download bandwidth \\
          $lat_{max}$: maximum admissible perception latency
\Ensure $(split, q)$: optimal parameter tuple
\end{algorithmic}
\setcounter{ALG@line}{0}
\begin{algorithmic}[1]
\Function{optPar}{$pArray$, $bw_{upl}$, $bw_{dwn}$, $lat_{max}$}
  \State $latMin \gets +\infty$
  \State $latMinIdx \gets 0$
  \For{$idx=0$ \textbf{to} $|pArray|-1$}
    \State $(split, q) \gets pArray[idx]$
    \State $lat \gets lat_{total}(split, q, bw_{upl}, bw_{dwn})$
    \If{$lat \leq lat_{max}$}
      \State \Return $(split, q)$ \Comment{\FH{highest-\ac{nds} feasible}}
    \ElsIf{$lat < latMin$} \Comment{\FH{lowest latency so far?}}
      \State $latMin \gets lat$
      \State $latMinIdx \gets idx$
    \EndIf
  \EndFor
  \State \Return $pArray[latMinIdx]$ \Comment{\FH{minimum-latency fallback}}
\EndFunction
\end{algorithmic}
\end{minipage}
}
\end{algorithm}

\textbf{Initialization:} At the start of the first iteration, the invariant holds trivially because no parameter tuples have been evaluated yet. 
The variables $latMin$ and $latMinIdx$ are initialized to a placeholder maximum value and index zero, respectively, ensuring that the first tuple's latency will always be accepted as the current minimum for comparison.

\textbf{Maintenance:} Assume the invariant holds at the beginning of iteration $idx$. 
\FH{After computing the latency $lat$ for tuple $pArray[idx]$,} if $lat < latMin$, then \FH{lines 10-11} update $latMin$ and $latMinIdx$ accordingly. Otherwise, these values remain unchanged. 
Therefore, \FH{at the start of the next iteration ($idx+1$),} $latMin$ and $latMinIdx$ still represent the minimum latency and corresponding index among all tuples examined so far\FH{. Hence,} the invariant holds.

\textbf{Termination:} If no latency-bound-satisfying tuple exists, the loop completes after examining all tuples. 
By the loop invariant, $latMinIdx$ then holds the index of the tuple with the minimum total latency. 
The algorithm returns this tuple in \FH{line 14}. 
This confirms the second part of Theorem~\ref{theo:optpar_correctness} and thus establishes the correctness of the algorithm.

The asymptotic running time of the algorithm depends on the input. 
Assuming that $lat_{total}$ can be computed in constant time, the \textbf{best case} occurs when the first tuple in the array satisfies the latency constraint. 
In this case, the loop terminates immediately, yielding a running time of $\mathcal{O}(1)$.
The \textbf{worst case} arises when no tuple satisfies the latency bound. 
The algorithm must then evaluate all entries in the array, resulting in a time complexity of $\mathcal{O}(|pArray|)$.
The \textbf{average case} depends on current network conditions and the distribution of latency values across tuples. 
\FH{If a valid configuration is typically found early within a constant number of iterations, the expected runtime approaches $\mathcal{O}(1)$. Otherwise, it becomes $\mathcal{O}(|pArray|)$.}

\subsection{Evaluation}
We \FH{now} describe how we evaluated the proposed dynamic parameter selection algorithm, and present the results.

\textbf{Setup:}
We performed a trace-based evaluation. The bandwidth dataset and source code used in this evaluation are publicly available on GitLab~\cite{DynParSelGitlab}.

\begin{figure}[!t]
    \centering
    \includegraphics[width=0.95\linewidth]{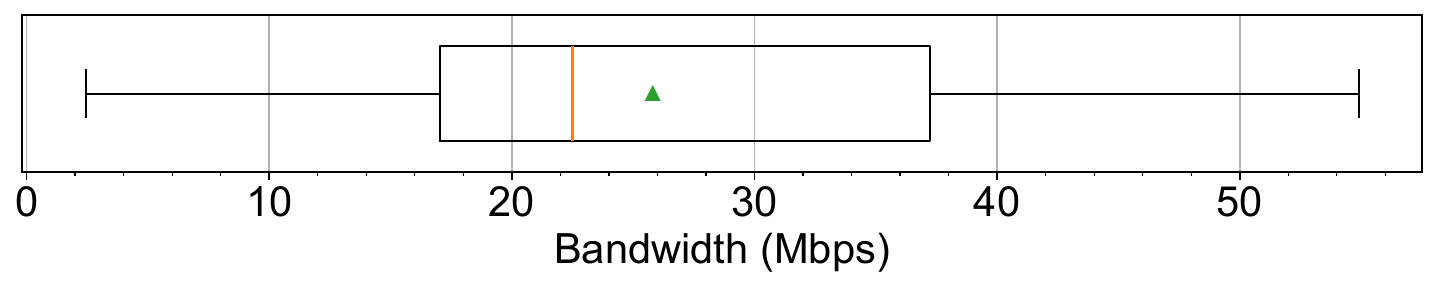}
    \caption{Upload bandwidth distribution used to evaluate the dynamic hybrid-computing parameter selection algorithm. The green triangle indicates the mean, and the orange line marks the median.}
    \label{fig:dyn-par-sel-dataset-bw-dist}
\end{figure}

We collected \SI{1}{\hertz} bandwidth measurements from a realistic vehicular scenario and used our algorithm to select the optimal split layer $split$ and quantization level for each entry. 
The process was repeated under varying latency limits $lat_{max}$ and bandwidth allocations for the perception task to assess performance under different constraints.
Because the dataset described in Section~\ref{subSec:test_route} does not feature direct bandwidth measurements, we collected a new one for this purpose. We used \texttt{iperf} running on a Raspberry Pi 4 Model B to stream UDP data to a cloud server, using a commercial 5G cellular network.
\FH{A total of \SI{654}{\second} of data were collected} from a vehicle traveling on an urban route in the city of Porto, Portugal\footnote{Bandwidth dataset collection route: \url{https://www.google.com/maps/d/viewer?mid=1ZpoUDMGTsPOjuKaYRngHe1botQZqyY8}}. 
Figure~\ref{fig:dyn-par-sel-dataset-bw-dist} depicts the distribution of measured bandwidths. The mean and standard deviation were \FH{\Mbitps{25.8}} and \Mbitps{12.1}, respectively.

Table~\ref{tab:delay_performance_breakdown} provided all additional information required to run the algorithm. 
Specifically, it supplied the \ac{nds} values used to rank parameter tuples by accuracy, as well as the data needed to compute the individual latency components in Equation~\ref{eq:latency_total}.

$lat_{local}$: The sum of the onboard processing time components associated with each parameter tuple.

$lat_{upl}$: The feature payload per perception cycle is obtained by converting the measured throughput at \SI{10}{\hertz} into bits per frame. The upload latency is then computed as the ratio between this payload and the available uplink bandwidth $bw_{upl}$.

$lat_{cloud}$: The sum of the cloud processing time components associated with each parameter tuple.

$lat_{down}$ is derived from the average C2V transmission latency reported in Table~\ref{tab:delay_performance_breakdown}, eliminating the need to rerun the experiments. 
This choice is justified by the empirical observation that C2V latency exhibits low standard deviation (e.g., $\pm 0.70$\,ms) and consistent average values across test runs, making it a reasonable proxy for downstream latency in our setting.
\begin{figure}[!t]
    \centering
    \includegraphics[width=0.99\linewidth]{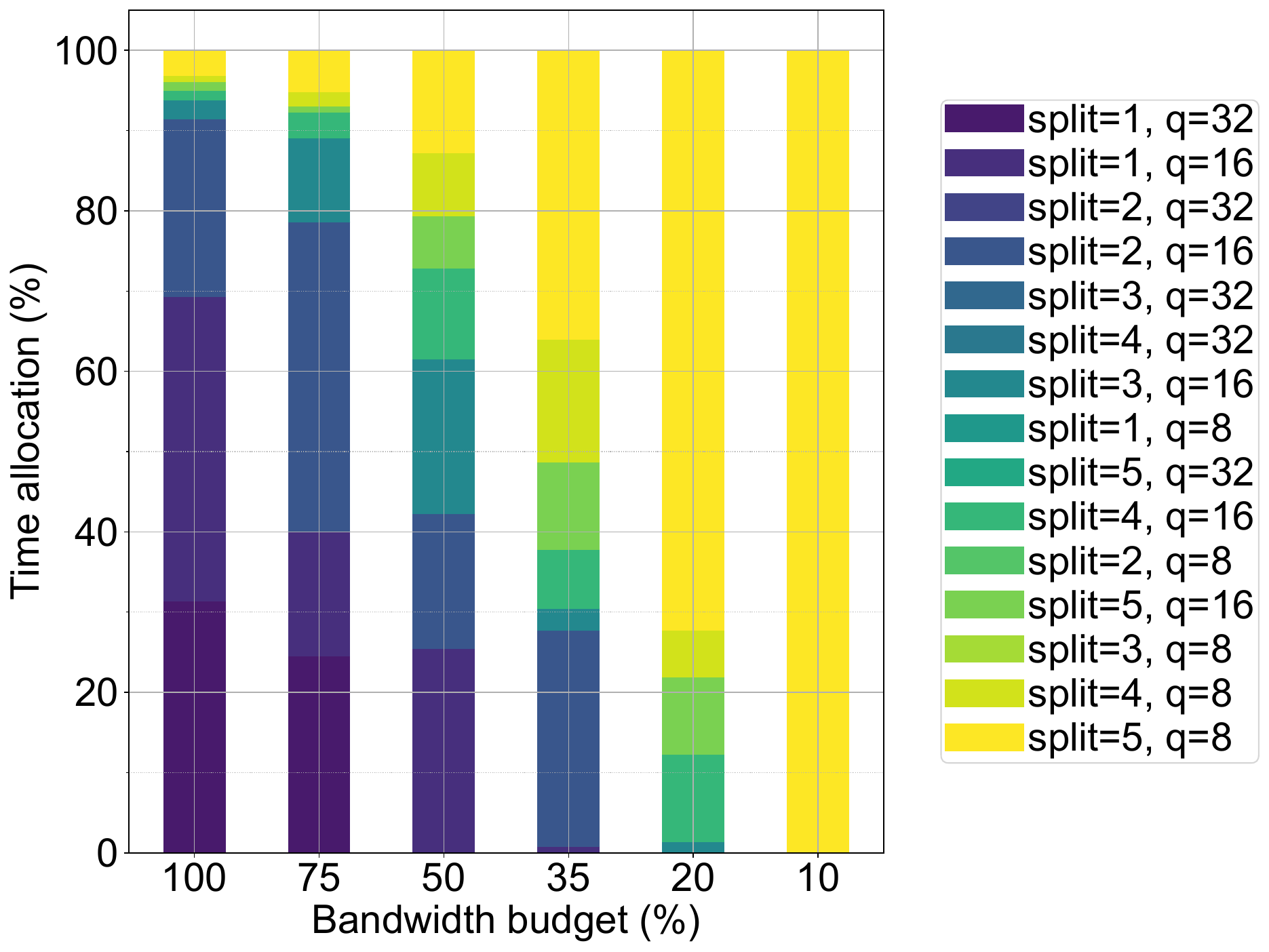}
    \caption{Distribution of parameter tuple usage by dynamic selection algorithm, as a function of the amount of bandwidth dedicated to the perception task.}
    \label{fig:dyn-par-sel-results-time-allocation}
\end{figure}

\subsection{Results Analysis}
We start by evaluating the dynamic algorithm's ability to adapt to different bandwidth budgets, with the maximum latency held constant at \ms{100}. 
Figure~\ref{fig:dyn-par-sel-results-time-allocation} shows the relative frequency with which each parameter tuple was selected, as a function of the bandwidth percentage dedicated to the perception task. 
\FH{As the available bandwidth increases, configurations yielding higher detection accuracy are selected more frequently.} 
When the budget was set to \pcent{100}, the two most-accurate tuples, (split=1, FP32) and (split=1, FP16), accounted for almost \pcent{70} of the total selections. 

\begin{figure}[!t]
    \centering
    \includegraphics[width=0.9209\linewidth]{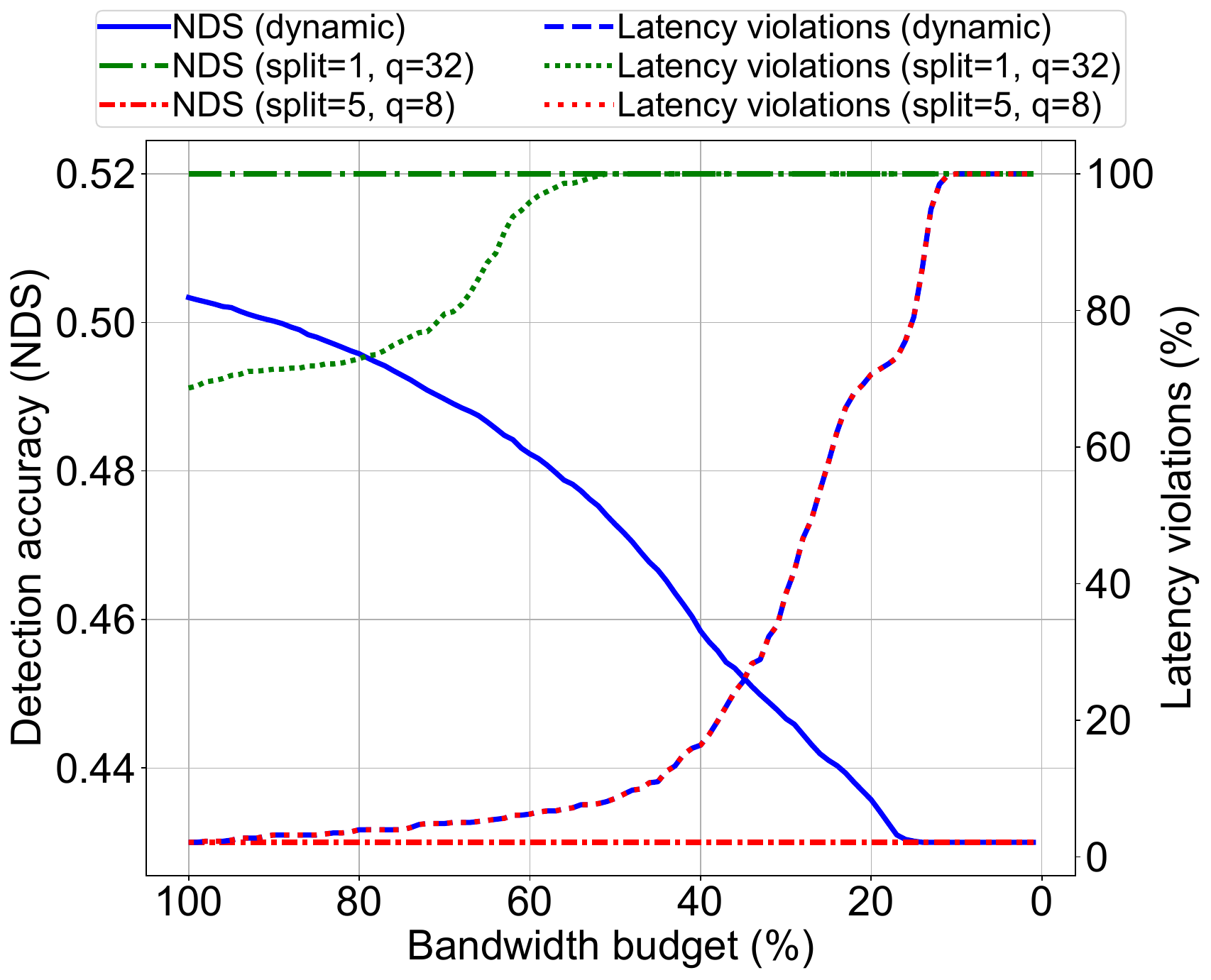}
    \caption{Detection accuracy (\ac{nds}) and number of latency bound violations, as a function of bandwidth dedicated to the perception task, for the dynamic parameter selection algorithm, the highest-accuracy static algorithm $(split=1, FP32)$, and the latency minimizing static algorithm $(split=5, FP8)$.}
    \label{fig:dyn-par-sel-results-nds-latvio-fo-bw}
\end{figure}

This dropped to slightly above \pcent{40} for a budget of \pcent{50}, and to zero for budgets of \pcent{20} or less. 
\FH{This behavior is consistent with the design of the algorithm:} decreasing the bandwidth budget increases communication latency, \FH{thereby forcing the optimizer to select lower-accuracy configurations} in order to satisfy the latency constraint.
Still holding maximum latency constant at \ms{100}, we now compare our dynamic algorithm's performance to that of the static ones, in terms of both detection accuracy, and number of latency bound violations. 
Figure~\ref{fig:dyn-par-sel-results-nds-latvio-fo-bw} plots these two metrics as a function of the bandwidth percentage assigned to the perception task. Three algorithms are represented: (i) the dynamic selection one, (ii) the accuracy-maximizing static configuration (split=1, FP32) and, (iii), the latency violations-minimizing static configuration (split=5, FP8).

We now examine detection accuracy (\ac{nds}). 
Because static configurations do not adapt to bandwidth variations, their accuracy remains constant: \SI{0.52}{} for (split=1, FP32) and \SI{0.43}{} for (split=5, FP8). 
In contrast, the dynamic algorithm's accuracy increases with available bandwidth, as higher bandwidth permits transmission of richer feature representations to the cloud. 
With a \pcent{100} bandwidth budget, the dynamic algorithm achieves accuracy within \pcent{5} of (split=1, FP32), \FH{while outperforming (split=5, FP8) by approximately \pcent{15}.
For a \pcent{50} bandwidth budget, the gap to (split=1, FP32) increases to roughly \pcent{10}, yet the dynamic method still outperforms (split=5, FP8) by about \pcent{10}.}
Overall, the dynamic algorithm preserves the latency robustness of the lowest-latency static configuration while providing a substantial improvement in detection accuracy.

\begin{figure}[!t]
    \centering
    \includegraphics[width=0.8154\linewidth]{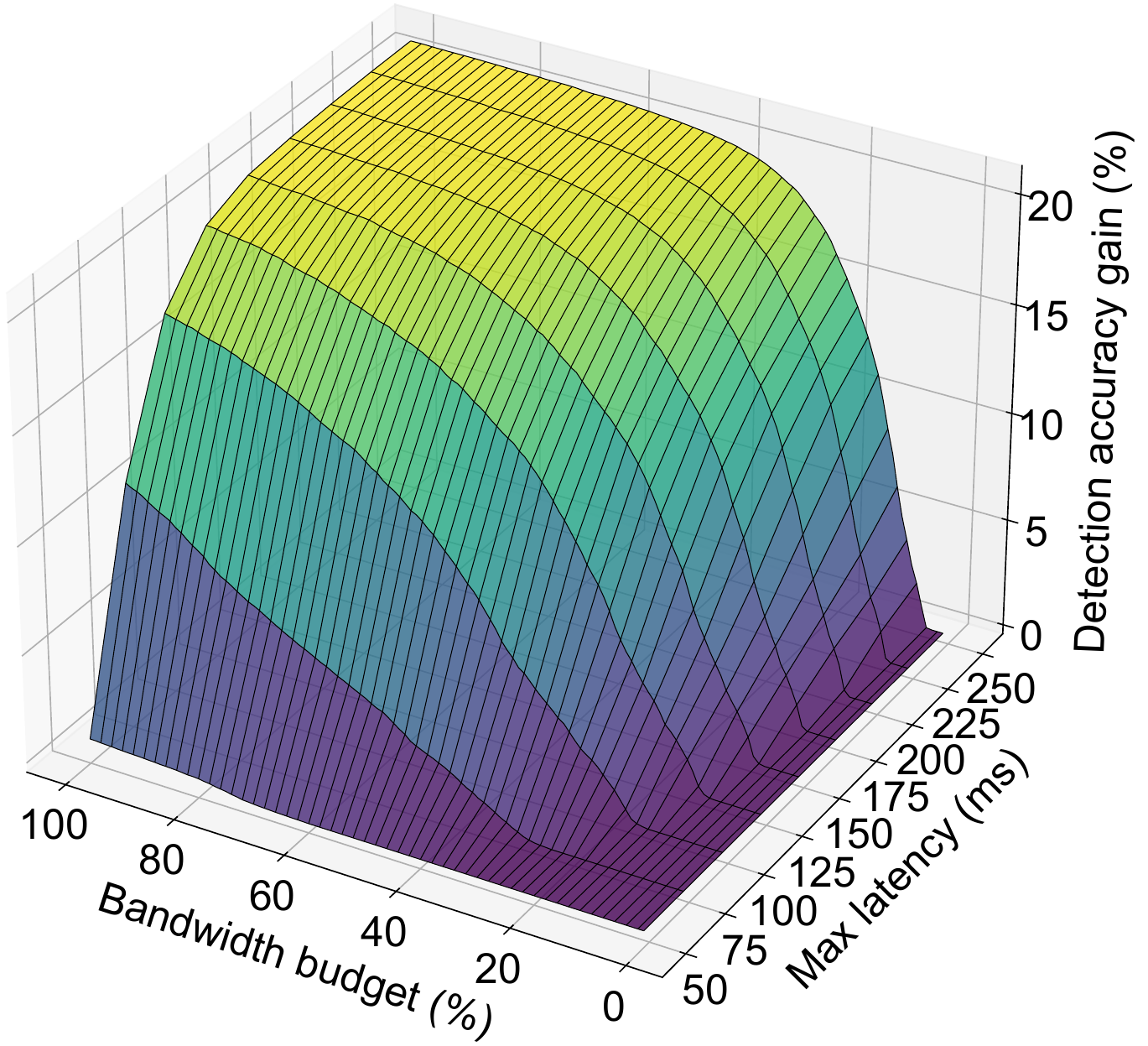}
    \caption{Detection accuracy (\ac{nds}) gain offered by dynamic selection algorithm, as a function of the amount of bandwidth assigned to the perception task.}
    \label{fig:dyn-par-sel-results-nds-gain-fo-bw-latmax}
\end{figure}

Finally, we explore the combined impact of latency limits and bandwidth budgets on detection accuracy. 
Figure~\ref{fig:dyn-par-sel-results-nds-gain-fo-bw-latmax} presents a surface plot of the accuracy gain achieved by the dynamic algorithm relative to the static configuration that minimizes latency violations for each $(lat_{max}, \text{bandwidth})$ pair. 
Across all evaluated scenarios, this baseline corresponds to the (split=5, FP8) configuration.

The curve corresponding to $lat_{max}=\SI{100}{ms}$ closely follows the trend observed in Figure~\ref{fig:dyn-par-sel-results-nds-latvio-fo-bw}. 
Reducing the latency bound further significantly decreases the achievable accuracy gains, which become nearly zero when $lat_{max}=\SI{50}{ms}$. 
This outcome is expected, since stricter latency constraints reduce the feasible set of high-accuracy configurations.
\FH{Conversely, relaxing the latency constraint enables the use of more data-intensive, higher-accuracy configurations, thereby increasing accuracy gains. 
For example, with a \pcent{100} bandwidth budget, increasing $lat_{max}$ from \SI{100}{ms} to \SI{250}{ms} raises the accuracy gain from \pcent{15} to \pcent{20}. 
Under a \pcent{50} bandwidth budget, the gain increases from \pcent{10} to \pcent{19}.} 
These results demonstrate the ability of the dynamic selection mechanism to adaptively optimize perception performance across diverse network conditions and latency requirements.

\FH{
\subsection{Model Choice and Generalizability}
\label{subsec:discussion_generalizability}
We employ BEVFormer with a ResNet101 backbone as a representative high-accuracy, compute-intensive camera-only BEV detector~\cite{li2022bevformer}. 
On the nuScenes benchmark, BEVFormer-ResNet101 achieves approximately 52 NDS depending on training configuration, while requiring substantial GPU memory and inference time. 
This makes it a suitable stress case for evaluating hybrid edge-cloud offloading under strict latency constraints.} 

\FH{Alternative BEV-style camera detectors (e.g., PETR~\cite{liu2022petr}, StreamPETR~\cite{wang2023exploring}) and different backbone depths exhibit distinct accuracy complexity trade-offs. 
For example, replacing ResNet101 with ResNet50 typically reduces FLOPs by approximately 20-30\% and decreases GPU memory usage accordingly, typically reducing NDS by a few points on nuScenes. 
Similarly, lighter BEV variants reduce intermediate feature dimensionality and inference latency at the cost of moderate performance degradation. 
In our adaptive framework, such changes directly affect (i) local computation time $lat_{local}$, (ii) transmitted feature size $S_{feat}$, and consequently (iii) the feasible $(split,q)$ configurations under a fixed latency constraint, as formalized in Eq.~\ref{eq:latency_total}. 
The proposed optimization framework is therefore not tied to a specific backbone, but rather to measurable accuracy, and latency trade-offs.}

\FH{Importantly, the proposed dynamic optimization framework is architecture-agnostic. 
It does not rely on model-specific structural assumptions. 
Instead, it operates on empirically profiled quantities for each candidate $(split,q)$ configuration: 
(i) detection accuracy (e.g., NDS), 
(ii) local computation time $lat_{local}(split,q)$, 
(iii) cloud computation time $lat_{cloud}(split,q)$, and 
(iv) transmitted feature size $S_{feat}(split,q)$. 
The latter determines the communication latency components under the available bandwidth. 
Given these profiles and the effective uplink/downlink transfer times, the optimizer selects the highest-accuracy configuration that satisfies the end-to-end latency constraint.
Consequently, the same methodology can be applied to alternative BEV detectors or backbone variants by re-profiling their split layers and quantization levels under the target hardware and network stack.}

\FH{Beyond backbone choice, the framework is compatible with alternative split learning and compression strategies. 
For instance, DeepSplit~\cite{mehta2020deepsplit} proposes learned intermediate-layer partitioning and compression policies for split inference. 
Such learned compression modules could replace the current quantization and compression pipeline. 
In contrast, our contribution focuses on bandwidth-adaptive selection of split depth and precision under strict latency constraints that explicitly account for network variability. 
Similarly, DistillBEV~\cite{wang2023distillbev} reduces computation and communication overhead through knowledge distillation, requiring additional training to obtain a compact model. 
The proposed optimizer remains directly applicable to such architectures, without structural modification, after profiling the corresponding $(split,q)$ configurations.}
\section{Conclusions and Future Work}
\label{sec:conclusion_future_work}
We proposed a hybrid-computing 360-degree 3D object detection system featuring a BEVFormer-based model coupled with V2X communication. 
The approach offloads intensive computations to the cloud while maintaining lightweight feature extraction onboard, enabling real-time perception.
Experimental results demonstrated that dynamic clipping, compression, and 5G-enabled C-V2X communication significantly optimize latency and bandwidth utilization. 
For instance, offloading FP32 feature vectors at \Hz{10} following a layer 1 split reduced the bandwidth requirement by approximately \pcent{98}, from \Mbitps{520} to \Mbitps{10.5}

We also investigated the trade-off between end-to-end delay and detection accuracy across various split layers and quantization levels. 

Our results demonstrated that while FP32 offers the highest accuracy, its substantial end-to-end delay renders it impractical for real-time applications. 
In contrast, FP8 achieved significantly lower latency with reasonable accuracy, making it suitable for latency-sensitive scenarios. 
FP16 provided a good compromise between accuracy and latency \FH{for applications that require both timely responses and adequate detection performance.}
This study underscores the importance of selecting appropriate split points and quantization levels based on operational requirements and conditions. 
Shallower splits with FP32 are optimal for accuracy-focused tasks, whereas deeper splits with FP8 cater to applications with strict latency constraints. 
\FH{Based on these findings,} we \FH{introduced} a dynamic parameter optimization algorithm that \FH{varies} the split layer and quantization level as a function of the available network bandwidth and target latency bound. 
In a trace-based evaluation, this algorithm matched the latency violation performance of the fastest static configuration while achieving double-digit accuracy gains across varying bandwidths and latency limits.
\FH{The achievable accuracy gain depends on the selected latency constraint. 
When the bound is strict, only a limited set of configurations remains feasible. 
When the bound is relaxed, higher-accuracy parameter combinations can be selected more frequently.}
\FH{The optimizer relies on empirically profiled latency and accuracy values. 
Therefore, it can be applied to other BEV backbones or compression strategies after profiling their configurations under the target hardware and network conditions.}

A promising direction is to integrate the dynamic parameter selection algorithm into a complete hybrid perception prototype and validate it in real-world driving scenarios\FH{, including multi-vehicle and dense traffic settings that exhibit stronger interference, higher contention, and rapid topology changes.}
This would enable end-to-end evaluation under realistic latency and bandwidth variations. 
Additionally, our results highlight the sensitivity of perception quality to bandwidth fluctuations, suggesting that combining the algorithm with 5G network slicing, specifically Ultra-Reliable Low Latency Communication (URLLC) slices could ensure consistent latency bounds and improve robustness in dynamic environments. 
Finally, exploring its applicability over emerging technologies such as 6G and satellite-based networks may extend its benefits to remote or under-connected regions, where stable communication is otherwise difficult to guarantee.
\section*{Acknowledgments}
This work is supported by the Fonds National de la Recherche of Luxembourg (FNR), under AFR grant agreement No 17020780 and project acronym \textit{ACDC}.
The authors would also like to thank Raquel Lopes from Instituto de Telecomunicações, for help collecting the vehicular cellular bandwidth dataset.
\balance
\bibliographystyle{elsarticle-num}
\bibliography{globo_refs}
\end{document}